\documentclass[lettersize,journal]{IEEEtran}
\usepackage{amsmath,amsfonts}
\usepackage{algorithmic}
\usepackage{algorithm}
\usepackage{array}
\usepackage[caption=false,font=normalsize,labelfont=sf,textfont=sf]{subfig}
\usepackage{textcomp}
\usepackage{stfloats}
\usepackage{url}
\usepackage{verbatim}
\usepackage{graphicx}
\usepackage{cite}
\usepackage{amsmath,amsfonts,bm}

\def\eqref#1{equation~\ref{#1}}

\def\1{\bm{1}}

\def\vf{{\bm{f}}}

\def\vs{{\bm{s}}}
\def\vt{{\bm{t}}}

\def\vv{{\bm{v}}}

\def\mA{{\bm{A}}}

\def\mE{{\bm{E}}}

\def\mI{{\bm{I}}}

\def\mO{{\bm{O}}}

\def\mQ{{\bm{Q}}}

\def\mV{{\bm{V}}}
\def\mW{{\bm{W}}}

\DeclareMathAlphabet{\mathsfit}{\encodingdefault}{\sfdefault}{m}{sl}
\SetMathAlphabet{\mathsfit}{bold}{\encodingdefault}{\sfdefault}{bx}{n}

\def\gN{{\mathcal{N}}}

\newcommand{\R}{\mathbb{R}}

\usepackage{algorithm}
\usepackage{algorithmic}

\usepackage{amsmath}
\usepackage{color}
\usepackage{multirow}

\newcommand{\ie}{\textit{i.e.}}
\newcommand{\eg}{\textit{e.g.}}
\newcommand{\etc}{\textit{etc}}

\makeatletter
\def\hlinewd#1{%
  \noalign{\ifnum0=`}\fi\hrule \@height #1 \futurelet
  \reserved@a\@xhline}
  
\begin{document}

\title{Keyword-Aware Relative Spatio-Temporal Graph Networks for Video Question Answering}


\author{Yi~Cheng,
        Hehe~Fan,
        Dongyun~Lin,
        Ying~Sun, \textit{Member, IEEE},
        Mohan~Kankanhalli, \textit{Fellow, IEEE},
        and Joo-Hwee~Lim, \textit{Senior Member, IEEE}

\thanks{Yi Cheng is with the Institute for Infocomm Research (I2R), Agency for Science, Technology and Research (A*STAR), and also with the School of Computing, National University of Singapore, Singapore (e-mail: cheng\_yi@i2r.a-star.edu.sg).}
\thanks{Dongyun Lin, Ying Sun are with the Institute for Infocomm Research (I2R), Agency for Science, Technology and Research (A*STAR), Singapore (e-mail: \{lin\_dongyun, liu\_yanzhu, suny\}@i2r.a-star.edu.sg).}
\thanks{Joo-Hwee Lim is with the Institute for Infocomm Research (I2R), Agency for Science, Technology and Research (A*STAR), Singapore, and also with the SCSE, Nanyang Technological University, Singapore (e-mail: joohwee@i2r.a-star.edu.sg).
}
\thanks{H. Fan and M. Kankanhalli are with the School of Computing, National University of Singapore, Singapore (e-mail: hehe.fan.cs@gmail.com, mohan@comp.nus.edu.sg).}
}

\markboth{Journal of \LaTeX\ Class Files,~Vol.~14, No.~8, August~2021}%
{Shell \MakeLowercase{\textit{et al.}}: A Sample Article Using IEEEtran.cls for IEEE Journals}


\maketitle

\begin{abstract}
The main challenge in video question answering (VideoQA) is to capture and understand the complex spatial and temporal relations between objects based on given questions. Existing graph-based methods for VideoQA usually ignore  keywords in questions and employ a simple graph to aggregate features without considering relative relations between objects, which may lead to inferior performance. In this paper, we propose a Keyword-aware Relative Spatio-Temporal (KRST) graph network for VideoQA. First, to make question features aware of keywords, we employ an attention mechanism to assign high weights to keywords during question encoding. The keyword-aware question features are then used to guide video graph construction. Second, because relations are relative, we integrate the relative relation modeling to better capture the spatio-temporal dynamics among object nodes. Moreover, we disentangle the spatio-temporal reasoning into an object-level spatial graph and a frame-level temporal graph, which reduces the impact of spatial and temporal relation reasoning on each other. Extensive experiments on the TGIF-QA, MSVD-QA and MSRVTT-QA datasets demonstrate the superiority of our KRST over multiple state-of-the-art methods. 
\end{abstract}


\section{Introduction}
\begin{figure}[h]
\begin{center}
  \includegraphics[width=1\linewidth]{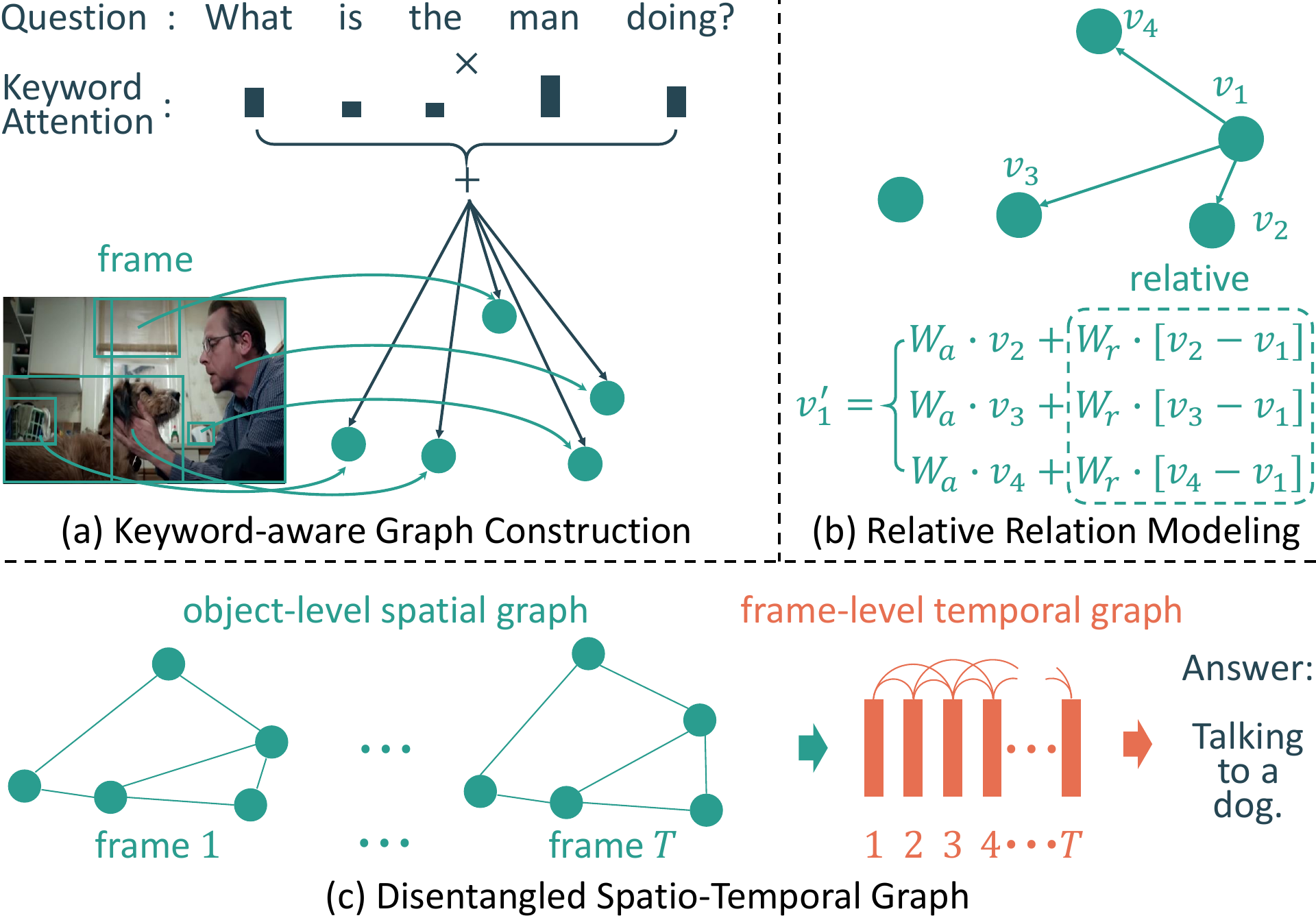}
\end{center}
\caption{Illustration of the proposed Keyword-aware  Relative Spatio-Temporal (KRST) graph network.
\textbf{(a)} We employ an attention mechanism to assign high weights to keywords during question embedding, which is then integrated into the object-level graph. 
\textbf{(b)} Relations are relative. Choosing different subjects (center objects) may lead to different understandings. For example, ``a man is \textit{talking} to a dog'' can also be understood as ``a dog is \textit{listening} to a man''. 
Therefore, we integrate the relative relation modeling into graph reasoning.
\textbf{(c)} Different from most existing methods that  employ a unified graph to model spatial-temporal relations together, we use an object-level graph for capturing spatial relation and a frame-level graph for temporal relation reasoning. 
} 
\label{fig:intro}
\end{figure}

\IEEEPARstart{V}{ideo} question answering (VideoQA) is a challenging task in Multimedia Intelligence~\cite{8970556, 10017364, 8768045}, and it aims to answer the question based on a thorough understanding of the given video.
The task requires the powerful cognitive capability of spatio-temporal visual representations guided by the compositional semantics of the given question. In recent years, VideoQA has drawn increasing attention due to its wide applications in various domains, \eg, human-robot interaction and autonomous driving. 

Despite its recent achievements, VideoQA still remains challenging as it requires effective reasoning about complex spatio-temporal relations based on the vision and language modalities~\cite{8811730, 9465732}.

To model the dynamics of relationships in videos, one solution is to capture the video structure at the frame level. 
For example, HCRN~\cite{hcrn} proposes a relation network to select relevant frames for each element in the context of the question query. 
Graph neural network (GNN) methods~\cite{jiang2020r, park2021b} are also used to model the temporal relations across frames by building graphs over video segments. 
However, because the object-level information is largely ignored, these frame-level methods are only able to model a limited number of objects and may fail to generalize to scenarios where multiple objects interact with each other. 

The second solution is to capture relations at the object level,  which is more flexible and able to model the complex spatial and temporal interactions among  more objects. 
To this end, GNNs are usually employed to model the relation structure among local objects~\cite{lgcn, gu2021graph, seo2021masn}.  
However, those methods treat all objects equally and potentially do not distinguish or recognize keywords in questions.
For the same video, different questions tend to focus on the interactions of different objects. 
Therefore, keyword-aware question embeddings should be exploited to guide the object-level graph construction. 
Moreover, those GNNs usually perform simple aggregation on neighbouring nodes (\eg, via an affinity matrix), in which the relative relations among objects are largely ignored. 
Without relative relation reasoning, networks may fail to properly capture the spatio-temporal structure. 

In this paper, we propose a Keyword-aware Relative Spatio-Temporal (KRST) graph network for VideoQA.
The graph is built over objects with guidance from questions.
First, as shown in Fig.~\ref{fig:intro}(a), to make question embeddings aware of keywords, we employ an attention mechanism to assign high weights to keywords during question encoding. The keyword-aware question features are then integrated into object features.
Our graph is based on those keyword-aware object representations.
Second, we integrate relative relation modeling into our graph. 
Our motivation is that relations are relative, whether for semantics or positions.  
For the same scene, choosing different subjects may lead to different relation understanding. 
As shown in Fig.~\ref{fig:intro}(b), when reasoning about the relation between the man and the  dog, it can be understood as ``a man is \textit{talking} to a dog'' if we focus on the man or ``a dog is \textit{listening} to a man'' if we select the dog as the target object: $man - dog = talk$ and $dog - man = listen$. 
Also, from the perspective of position, the scene can be understood as ``the dog is left to the man'' or ``the man is right to the dog'': $man - dog = right$ and $dog - man = left$. 
Therefore, it is important to equip graph reasoning with the relative relation modeling ability.
Third, instead of modeling spatio-temporal relations jointly, we disentangle the modeling into an object-level graph for spatial reasoning and a frame-level graph for temporal structure capture, as shown in Fig.~\ref{fig:intro}(c). In this way, the object graph only focuses on extracting the object spatial relations of interest regarding the question while the frame-level graph only needs to capture the dynamics of the attended object relations, which reduces the burden of networks for reasoning.   
The contributions of this paper are threefold:
\begin{itemize} 
  \item We propose a Keyword-aware Relative Spatio-Temporal (KRST) graph network for VideoQA, which performs relation reasoning over objects with keyword-aware question features. 
  \item We introduce relative relation modeling for VideoQA, which enables graph networks to better understand spatial-temporal relationships among objects. 
  \item We conduct extensive experiments on TGIF-QA, MSVD-QA and MSRVTT-QA datasets, and the results demonstrate the superiority of our KRST to the state-of-the-arts.
\end{itemize}

\section{Related Work}
Recently, there has been a rapid progress in vision-language tasks~\cite{li2022fine,zhang2021DVC, fan2020recurrent, zhu2020actbert, wang2021t2vlad, zhao2022centerclip,zhu2017uncovering, DBLP:journals/tip/YangWDDWC22}, such as image captioning, visual question answering, visual dialog, text-video retrieval, \etc. According to the types of visual information, question answering (QA) can be classified into image QA and video QA. Compared with image QA, video QA is more challenging because both spatial and temporal relations are required to be modelled for correct answer prediction~\cite{zhong2022video}. 

Inspired by the recent advances in large vision and language models, some existing VideoQA methods~\cite{yang2021just, zellers2021merlot, yu2021learning, Li_2022_CVPR, miech2019howto100m} attempt to extract richer information from videos and questions by applying large-scale pre-trained models. However, this work focuses on exploiting the interactions between semantic clues from the visual contents and linguistic questions. 
Existing methods on VideoQA can be grouped into two categories: attention-based and graph-based methods.

\paragraph{Attention-based methods}
Typical attention-based methods~\cite{tvqa,li19,jin19,8811730} learn temporal attention by modeling the correlation between appearance and question. For example, co-attention ~\cite{li19} is proposed to model appearance and question interaction, while object-level attention~\cite{jin19} is proposed to learn object-question interaction. 
Some other works learn spatio-temporal attention by leveraging both appearance and motion features, including co-memory attention~\cite{comem}, hierarchical attention~\cite{zhao18}, multi-head attention~\cite{kim18} and multi-step progressive attention~\cite{msvd-qa}.
HCRN~\cite{hcrn} proposes a hierarchical structure to model the  temporal relation of a single object with a stack of conditional relation blocks. 
However, these methods may fail to handle cases where multiple objects interact. 

\paragraph{Graph-based methods}
Graph convolutional network (GCN) has been widely applied to various vision-language tasks~\cite{li2022compositional, nan2021interventional,li2021adaptive}.
For VideoQA, L-GCN~\cite{lgcn} proposes a location-aware graph to model the relation between different objects in each video frame. 
GMN~\cite{gu2021graph} designs a holistic spatio-temporal graph to model the relations between different objects  by taking object trajectories as input. 
MASN~\cite{seo2021masn} extends the spatio-temporal graph over objects by exploiting the complementary nature between appearance and motion information.
HOSTR~\cite{da2021hrac} leverages the question features to compute the correlation matrix between objects for relation reasoning. 
HGA~\cite{jiang2020r} proposes a uniform heterogeneous graph with video shots and words as nodes to incorporate both inter- and intra-modality interactions. B2A~\cite{park2021b} constructs multi-modal graphs for appearance, motion and questions, where the question graph is used  to model the relation between video and question. 

\begin{figure*}[t]
\begin{center}
   \includegraphics[width=1\linewidth]{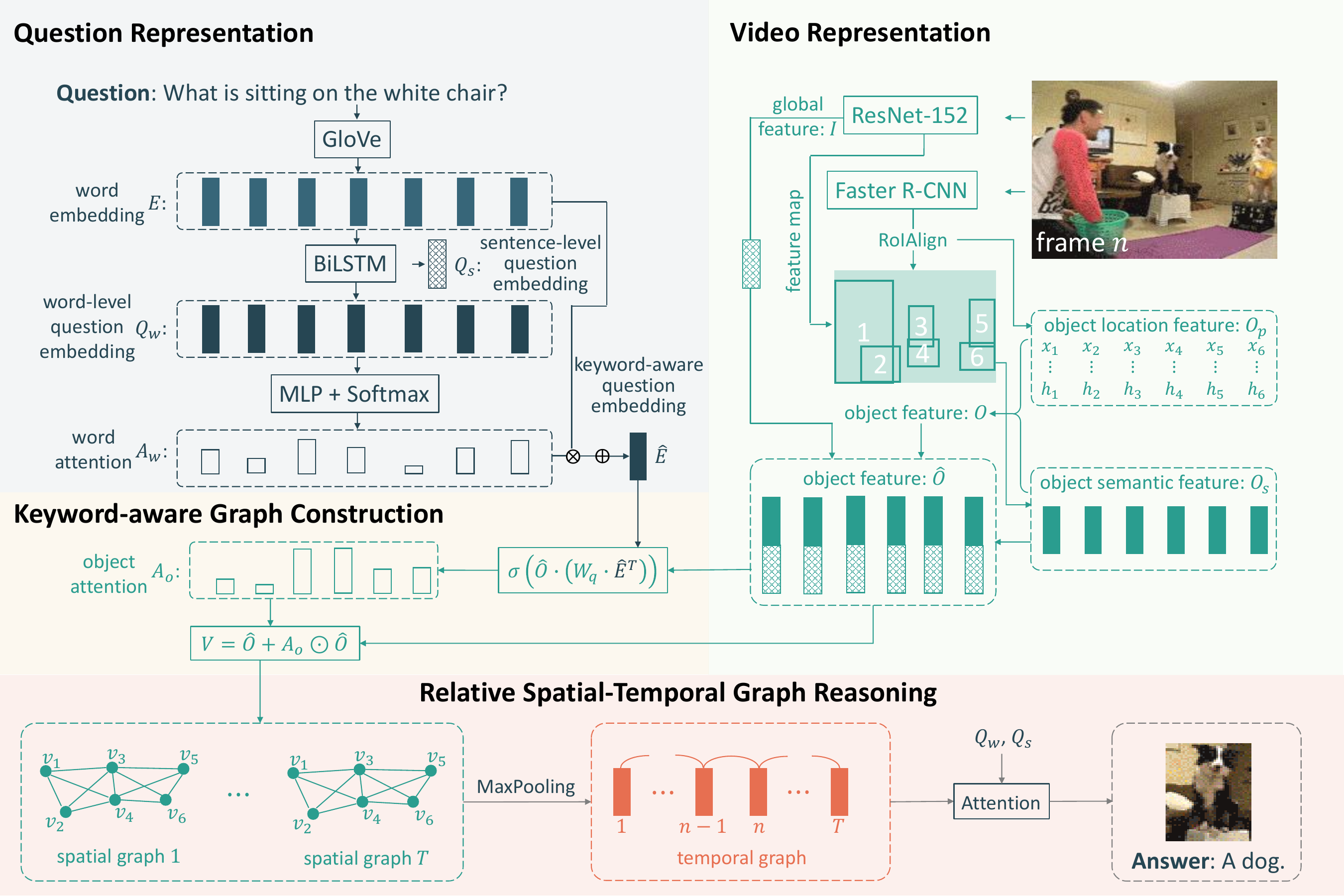}
\end{center}
\caption{
The architecture of the proposed Keyword-Aware Relative Spatio-Temporal graph network for VideoQA.
The video and question features are firstly extracted as described in Section~\ref{method:fea_gen}.
Then, then we construct the keyword-aware graph by designing a keyword attention module to identify and
augment relevant object features for the subsequent relation reasoning. 
Next, we perform relative spatio-temporal graph reasoning over objects using disentangled spatial and temporal graphs.
Finally, the question features and visual features are fused to predict the final answer.
}
\label{fig:1}
\end{figure*}

Different from these approaches, our method builds spatio-temporal graphs by attending to keyword-relevant objects. 
Moreover, we integrate relative relation modeling into spatio-temporal graph reasoning. 
Last, to explicitly model the spatial and temporal relations in videos, we decompose the spatio-temporal graphs into a spatial graph over objects within each frame and a temporal graph across frames.

\section{Methodology}

Given a video $\mathcal{I}$ and a linguistic question $q$, the VideoQA problem can be formulated as follows, 
\begin{equation}
\bar{a}=\arg\max_{a\in \mathcal{A}}\mathcal{F}_{\theta}\left(a \mid q, \mathcal{I}\right),
\label{eq:VQA-overall}
\end{equation}
where $\bar{a}$ is the predicted answer from an answer set $\mathcal{A}$. $\mathcal{F}_{\theta}(.)$ is the mapping function and $\theta$ is the set of parameters. 
Fig.~\ref{fig:1} illustrates the overall architecture of the proposed KRST method.
First, the question embeddings are extracted via GloVe~\cite{pennington2014glove} and Bidirectional LSTM (BiLSTM).
We design a keyword-aware module to guide question embedding, which promotes a deep understanding of object interactions by reducing the noise from question-irrelevant objects.
Second, the object-level features are generated using pre-trained models based on RoIAlign. 
Those object-level features are then used to build the graph, guided by the keyword-aware question embedding. 
We integrate the relative relation modeling into graph reasoning. 
Finally, both object- and frame-level features are fused with question features to predict the final answer. 

\subsection{Video and Question Representation}
\label{method:fea_gen}

\subsubsection{Video-level Appearance and Motion Representation}
We follow the common practice in previous works to extract both appearance and motion features from videos. 
Specifically, $T$ frames are uniformly sampled from each video. Then, pre-trained ResNet-152~\cite{he2016deep} and I3D~\cite{carreira2017quo} models are employed to extract the global context appearance feature and motion feature, respectively.
For motion feature extraction, because I3D takes multiple frames as input, a set of 8 neighboring frames around each sampled frame is concatenated and fed into I3D model. 
Based on these two types of features, we build a two-stream architecture~\cite{DBLP:conf/nips/SimonyanZ14}. 

Note that, for simplicity, in the following descriptions, we do not make notes or subscripts to distinguish appearance and motion features but treat them as visual features. 
In this case, we let $\mI \in \R^{T \times C}$ denote the video-level appearance or motion features, where $C$ denotes the feature dimension. 

\subsubsection{Object-level Appearance and Motion Representation}
To obtain object-level features, we first generate $K$ object bounding boxes from each frame by employing a pre-trained object detector Faster R-CNN~\cite{ren2016faster}. In this case, there are $TK$ objects in each video. Then, the object semantic features $\mO_s \in \R^{TK \times C_s}$ for appearance or motion are obtained by applying RoIAlign onto the video-level ResNet-152 or  I3D feature maps, respectively, where $C_s$ is the dimension of the semantic feature maps.  
To effectively model object relations, it is important to integrate the object location information into object features. 
In this paper, we use the $x$ and $y$ coordinates of the upper-left corner, such coordinates of the lower-right corner and the height and width of the bounding box, \ie, $(x_1, y_1, x_2, y_2, w, h)$,  as the object location information: $\mO_p \in \R^{TK\times6}$.
Similar to L-GCN~\cite{lgcn}, we first concatenate the object semantic features and the  bounding box  location. 
Then, the object features are obtained by projecting the concatenated features into $C$-dimension using a linear transformation,
\begin{equation}
\label{eq:position}
    \mO = \mW_o \cdot [\mO_s, \mO_p], 
\end{equation}
where $\mW_o \in \R^{C \times (C_s + 6)}$, $\cdot$ is matrix multiplication and $[\cdot,\cdot]$ denotes the concatenation operation along rows.

\subsubsection{Video-level and Object-level Representation Fusion}
To capture the background contextual information and compensate for potentially undetected objects, we augment the object features with video-level features. 
The final object-level features $\hat{\mO} \in \R^{TK \times C_o}$ are generated as follows, 
\begin{equation}
\label{eq:locenc}
    \hat{\mO} = \mathrm{MLP} \big([\mO, \mathrm{tile}(\mI)] \big),
\end{equation}
where $\mathrm{tile}(\cdot)$ repeats $\mI$ by $K$ times to $\R^{TK\times C}$ and $C_o$ is the dimension of object node features. 
MLP is a multilayer perception to project $[\mO, \mathrm{tile}(\mI)]$ from $\R^{TK \times 2C}$ to $\R^{TK \times C_o}$. 

\subsubsection{Question Representation.}
Given a question, we first embed its words into vectors $\bm{E} \in \mathbb{R}^{L \times 300}$ using a pre-trained GloVe, where $L$ denotes the number of words in the question. Then, we feed the embedded vectors into a BiLSTM to generate a contextualized word-level question embedding $\mQ_w \in \R^{L \times C_w}$, where $C_w$ is the question feature dimension, and a global or sentence-level question embedding $\mQ_s \in \R^{1 \times C_w}$, which is the last hidden state of the BiLSTM.

\subsection{Keyword-aware Graph Construction}
To correctly answer the question based on a given video, the agent is required to reason about the complex interactions between objects. As videos usually contain multiple objects, it is important to attend to question-relevant objects, which can be seen as keywords, without having the noise from irrelevant objects.
However, most existing VideoQA methods with object-level relation modeling treat all the objects equally without considering their relevance to the keywords. Consequently, the redundant information from irrelevant objects may reduce reasoning accuracy. To overcome this limitation, we design a keyword attention module to identify and augment relevant object features for the subsequent relation reasoning. 
Specifically, we first compute the attended question feature $\hat{\bm{E}} \in \mathbb{R}^{1 \times 300}$ with the attention mechanism, 
\begin{equation}
    \mA_w = \mathrm{softmax}(\mathrm{MLP}(\mQ_w)), \ \ \ \ \hat{\mE} = \mA_w^T \cdot \mE, 
    \label{eq:word_attn}
\end{equation}
where $\mA_w \in \R^{L \times 1}$ is the attention weights. 
Then, we can use the attended question feature to guide the generation of object attention,
\begin{equation}
    \mA_o = \sigma \big(\hat{\mO} \cdot (\mW_q \cdot  \hat{\mE}^T) \big), 
    \label{eq:object_attn}
\end{equation}
where  $\sigma$ is the sigmoid function, $\mW_q \in \R^{C_o \times 300}$ and $\mA_o \in \R^{TK \times 1}$. 
Thus, the keyword-relevant objects are highlighted by assigning higher attention scores. Consequently, these objects will be attended in relation reasoning.
Finally, the object node features are generated as follows,
\begin{equation}
    \mV = \hat{\mO} + \mA_o \odot \hat{\mO},
\end{equation}
where $\odot$ denotes the Hadamard product and $\mV \in \R^{TK \times C_o}$.

\subsection{Relative Spatial-Temporal Graph Reasoning}
Most existing VideoQA methods model the complex object relations by representing a video as a holistic spatio-temporal graph over all the detected object proposals~\cite{gu2021graph,seo2021masn}. 
These methods usually simply aggregate neighboring nodes with affinity metrics, in which the relative relation information is largely ignored. 
However, relations are relative. 
Choosing different objects of interest may lead to different semantic and position relation reasoning. 
Taking the example of a writer writing a book, if the writer is chosen as the subject and the book as the object, the relation is ``write''. 
Conversely, the relation is ``author'', \ie, the book's author is the writer.
Similarly, for the position relation, a cup on the table can also be understood as the table  under the cup.
Therefore, we integrate the relative relation modeling into our graph.

Let $\vv_t^i \in \R^{1 \times C_o}$ denote the feature of the $i$-th object or node in the $t$-th frame of $\mV$. 
Our relative-relation-augmented graph models the relations between the object and its neighbors as follows,
\begin{equation}
\label{eq:relative}
   {\vv'}_t^i = \max_{\vv_{t'}^{i'} \in \gN(\vv_t^i)} \mW_a \cdot \vv_{t'}^{i'} + \mW_r \cdot (\vv_{t'}^{i'} - \vv_t^i), 
\end{equation}
where $\gN(\vv_t^i)$ denotes the k-nearest neighbors of $\vv_t^i$ in the representation space, $\mW_a \in \R^{C \times C_o}$ is for absolute relation modeling and $\mW_r \in \R^{C \times C_o}$ is for relative relation reasoning. 
Thus, our graph will be able to realize the relativity in relations.

\subsection{Disentangled Spatial-Temporal Graph}

Spatial and temporal relations are two different types of relations.
Modeling them together may confuse networks and reduce the reasoning efficacy. 
Moreover, representing a video as a graph over all the objects is computationally expensive.
Efficient information flow in such a large graph is challenging.
Therefore, we propose to disentangle the graph in Eq.~(\ref{eq:relative}) into a spatial graph and a temporal graph,
\begin{equation}
\small
    \begin{split}
        & \mathrm{spatial}:  \ \  \ \ \ \ \  \vs_t^i = \max_{\vv_t^{i'} \in \gN(\vv_t^i)} \mW_a^s \cdot  \vv_t^{i'} + \mW_r^s \cdot (\vv_t^{i'} - \vv_t^i), \\
        & \mathrm{aggregation}:     {\vf}_t =  \max_{i=1}^K {\vs}_t^i,  \\     
        & \mathrm{temporal}: \ \ \ \ \  {\vt}_t = \sum_{\vf_{t'} \in \gN(\vf_t)} \mW_a^t \cdot \vf_{t'} + \mW_r^t \cdot (\vf_{t'} - \vf_t), \\
    \end{split}
\label{eq:aggr}
\end{equation}
where $\mW_a^s, \mW_r^s,\mW_a^t, \mW_r^t \in \R^{C \times C_o}$ are the parameters for spatial and temporal reasoning.

In the spatial reasoning and aggregation parts, we use the $\max$ pooling operation to keep the most relevant nodes to the question and ignore the noise from the other irrelevant objects. 
In the temporal reasoning part, we employ the sum pooling operation. 
This is because in a short video clip, the object of interest usually appears in the entire video and the spatial relation in each frame is important to the overall dynamics reasoning. 
As shown in experiments, this max-sum pooling combination can achieve better accuracy than others, also demonstrating the benefits of this disentangled spatio-temporal modeling. 

To generate the final output vector for answer prediction, we use bilinear attention \cite{kim2018bilinear} to project spatial graph presentations $(\{\vs_t^i\}, \mQ_w)$ and temporal graph presentations $(\{\vt_t\}, \mQ_w)$ to the same space as the question word-level features $\mQ_w$'s $\R^{L \times C_w}$, respectively. 
Then, we employ the attention mechanism in~\cite{seo2021masn} to fuse the projected spatial features, temporal features and the question sentence-level features $\mQ_s$ to a vector, which is finally used to answer the question.

\subsection{Answer Decoder}
VideoQA generally includes two types of tasks: multi-choice and open-ended. In this section, we design the answer decoders to deal with different tasks. 
The multi-choice task aims to choose the correct answer from $M$ candidates. In this case, we concatenate the question with each answer and obtain $M$ question-answer sequences. For each sequence, we feed it into the network to compute the final output vector, and then employ a fully-connected layer to generate the predicted score $a_i$. Suppose $a^+$ is the score of the correct answer and $(a_1^-, \dots, a_{M -1}^-)$ are the scores of wrong answers. We use the pairwise hinge loss $\sum_{i=1}^{M-1}\max \big(0, 1 - (a^+ - a_i^-)\big)$ to train networks. 

The open-ended task aims to choose the correct answer from a pre-defined answer set, which can be treated as a multi-label classification task. Therefore, we feed the output of our GNN into a classifier with two fully-connected layers to compute the class probabilities and train the network using the cross-entropy loss. For counting, \ie, open-ended numbers, we treat it as a regression task and train the network using the Mean Squared Error (MSE) loss. 

\section{Experiments}

\begin{table}[t]
\renewcommand\arraystretch{1.1}
\caption{Results on the TGIF-QA dataset. GloVe is used for word embedding.}
\begin{center}
\setlength{\tabcolsep}{1mm}{
\scalebox{1.0}{
    \begin{tabular}{l c c c c }
    \hlinewd{0.8pt}
         Model & Action$\uparrow$ & Transition$\uparrow$ & Frame$\uparrow$ & Count$\downarrow$  \\
        \hline \hline
        ST-TP~\cite{tgif-qa}    &  62.9    &   69.4    &   49.5    &   4.32   \\
        Co-mem~\cite{comem}     &   68.2    &   74.3    &   51.5    &   4.10   \\
        PSAC~\cite{psac}        &   70.4    &   76.9    &   55.7    &   4.27    \\
        QueST~\cite{quest}      &   75.9    &   81.0    &   59.7    &   4.19  \\
        HCRN~\cite{hcrn}        &   75.0    &   81.4    &   55.9    &   3.82     \\   
        \hline
        HGA~\cite{jiang2020r}   &   75.4    &   81.0    &   55.1    &   4.09  \\
        B2A~\cite{park2021b}    &   75.9    &   82.6    &   57.5    &   3.71  \\
        \hline
        L-GCN~\cite{lgcn}       &   74.3    &   81.1    &   56.3    &   3.95  \\
        GMN~\cite{gu2021graph}  &   73.0    &   81.7    &   57.5    &   4.16  \\
        HOSTR~\cite{da2021hrac} &   75.0    &   83.0    &   58.0    &   3.65  \\
        MASN~\cite{seo2021masn} &   84.4    &   87.4    &   59.5    &   3.75  \\
        HQGA~\cite{xiao2022video} &   76.9    &   85.6    &   \textbf{61.3}    &   -  \\
        \hline
        KRST (ours)   &\textbf{85.0}   &\textbf{88.8}  & 60.9  &  \textbf{3.62}  \\
    \hlinewd{0.8pt}
    \end{tabular}}}
\end{center}
\label{table:sota_tgifqa}
\end{table}

\subsection{Datasets}
We evaluate the proposed method on the three most commonly used benchmarks for VideoQA task.
\paragraph{TGIF-QA} It is a large-scale VideoQA dataset with $165K$ Q\&A pairs from $72K$ animated GIF videos~\cite{tgif-qa}. There are four types of questions in this dataset: 1) \textit{Action}: multiple-choice questions to identify the action repeated for certain times; 2) \textit{Transition}: multiple-choice questions to identify the action regarding state transition; 3) \textit{FrameQA}: open-ended questions that can be inferred from one frame in a video; 4) \textit{Count}: open-ended questions to count the number of occurrences of an action. Multiple-choice questions have five options, while open-ended questions have a pre-defined answer set of size $1,746$.

\paragraph{MSVD-QA \& MSRVTT-QA} Both datasets contain $50K$ Q\&A pairs from 2K short videos~\cite{msvd-qa} and $243K$ Q\&A pairs from 10K short videos~\cite{msrvtt-qa}, respectively. They cover five different question types: what, who, how, when and where. The questions are open-ended with a pre-defined answer set of size $1K$ and $4K$, respectively.

\paragraph{Evaluation Metrics}
We use MSE for the \textit{Count} task in TGIF-QA and accuracy for  other tasks.

\subsection{Implementation Details}
We extract videos at $10$ frames per second for all the datasets, and uniformly sample $T$ frames from each video. According to the average video length, we set $T$ as $30$ for MSVD-QA \& MSRVTT-QA datasets and $20$ for TGIF-QA dataset. 
For the global context feature extraction, we apply ResNet-152 model pretrained on the ImageNet~\cite{deng2009imagenet} dataset for appearance features and I3D model pre-trained on the Kinetics~\cite{carreira2017quo} dataset for motion features. 
For the local feature extraction, we select $K=10$ object bounding boxes with the highest confidence scores for each frame, where the bounding boxes are generated using Faster R-CNN~\cite{ren2016faster} pretrained on the Visual Genome~\cite{krishna2017visual} dataset. We set the number of graph layers $H$ in both spatial and temporal graphs as $2$ to capture spatial and temporal information. The ratio $\alpha$ to control number of neighboring nodes in spatial and temporal graphs is set as $0.6$ and $0.8$, respectively.
Moreover, the model's hidden size $d$ is set as $512$. 

The proposed model is trained on PyTorch~\cite{paszke2019pytorch}. During training, we set batch size as $64$ for multi-choice questions and $128$ for open-ended and count questions. The dropout ratio is set as $0.3$ to prevent overfitting. We train the model for $30$ epochs using the Adam optimizer with a constant learning rate of $10^{-4}$. Experiments are implemented on Ubuntu
20.04 by NVIDIA GTX 3090 GPUs.

\begin{table}[t]
\renewcommand\arraystretch{1.1}
\caption{Results on the MSVD-QA and MSRVTT-QA datasets. GloVe is used for word embedding.}
\begin{center}
\scalebox{1.0}{
    \begin{tabular}{l c c }
    \hlinewd{0.8pt}
        Model                           & MSVD-QA & MSRVTT-QA \\
        \hline \hline
        AMU~\cite{msvd-qa}              &  32.0     &   32.5    \\              
        ST-TP~\cite{tgif-qa}            &  30.9     &   31.3    \\
        Co-mem~\cite{comem}             &   31.7    &   31.9    \\
        QueST~\cite{quest}              &   36.1    &   34.6    \\
        HCRN~\cite{hcrn}                &   36.1    &   35.6    \\ 
        
        \hline
        HGA~\cite{jiang2020r}           &   34.7    & 35.5 \\
        B2A~\cite{park2021b}            &   37.2    & 36.9 \\
        \hline
        L-GCN~\cite{lgcn}               &   34.3    &   -    \\
        GMN~\cite{gu2021graph}          &   35.4    & 36.1 \\
        HOSTR~\cite{da2021hrac}         &   39.4    & 35.9 \\
        MASN~\cite{seo2021masn}         &   38.0    &   35.2    \\ 
        HQGA~\cite{xiao2022video}       &   41.2    &   37.2    \\
        \hline
        KRST (ours)                   &\textbf{41.5}   &\textbf{37.4}   \\
    \hlinewd{0.8pt}
    \end{tabular}}
\end{center}
\label{table:sota_msvdmsrrt}
\end{table} 

\subsection{Comparison with the State-of-the-art}
Table~\ref{table:sota_tgifqa} and Table~\ref{table:sota_msvdmsrrt} present the performance comparison of our KRST with multiple SOTA methods on three popular benchmarks. The comparing methods can be classified into three categories: \textbf{(1) attention-based methods}, including ST-TP~\cite{tgif-qa}, Co-mem~\cite{comem}, PSAC~\cite{psac}, QueST~\cite{quest} and HCRN~\cite{hcrn}; \textbf{(2) graph-based methods on frame relations}, including HGA~\cite{jiang2020r} and B2A~\cite{park2021b}; \textbf{(3) graph-based methods on object relations}, including L-GCN~\cite{lgcn}, GMN~\cite{gu2021graph}, HOSTR~\cite{da2021hrac}, MASN~\cite{seo2021masn} and HQGA~\cite{xiao2022video}.
The experimental results demonstrate that our model consistently outperforms the SOTA models on three datasets.
Note that it is challenging to achieve the best accuracies on all of the three datasets. 
For example, MASN~\cite{seo2021masn} achieves quite a high accuracy on the TGIF-QA dataset but only reaches inferior accuracies on the MSVD-QA and MSRVTT-QA datasets.
Similarly, HQGA~\cite{xiao2022video} can achieve satisfactory performance on MSVD-QA and MSRVTT-QA but not on TGIF-QA (especially for the action task).
In contrast, experiments show the superiority of our method on the three datasets, which implies that our method can effectively capture complex object interactions in space and time for VideoQA.

HGA and B2A are graph-based methods that model the temporal relations between different frames. In these methods, multiple graphs are constructed to exploit the correlation between question words and video segments. However, they ignore the fine-grained object-level information important for question answering, leading to inferior performance.

GMN and HOSTR build a holistic spatio-temporal graph to model object relations by taking object trajectories as input. They assume objects exist throughout the video, which may be invalid for videos with multiple objects and heavy occlusions. This assumption may bring extra noise in object relation modeling and result in inferior performance. 
MASN and HQGA are similar to our work in taking independent object proposals as input. 
However, both MASN and HQGA perform relation reasoning ignoring the keywords from questions, and they do not consider the relative relation between objects.
By filling these gaps, our model demonstrates a clear superiority over these SOTA methods on all the three datasets.

\begin{figure}[t]
\begin{center}
  \includegraphics[width=1.0\linewidth]{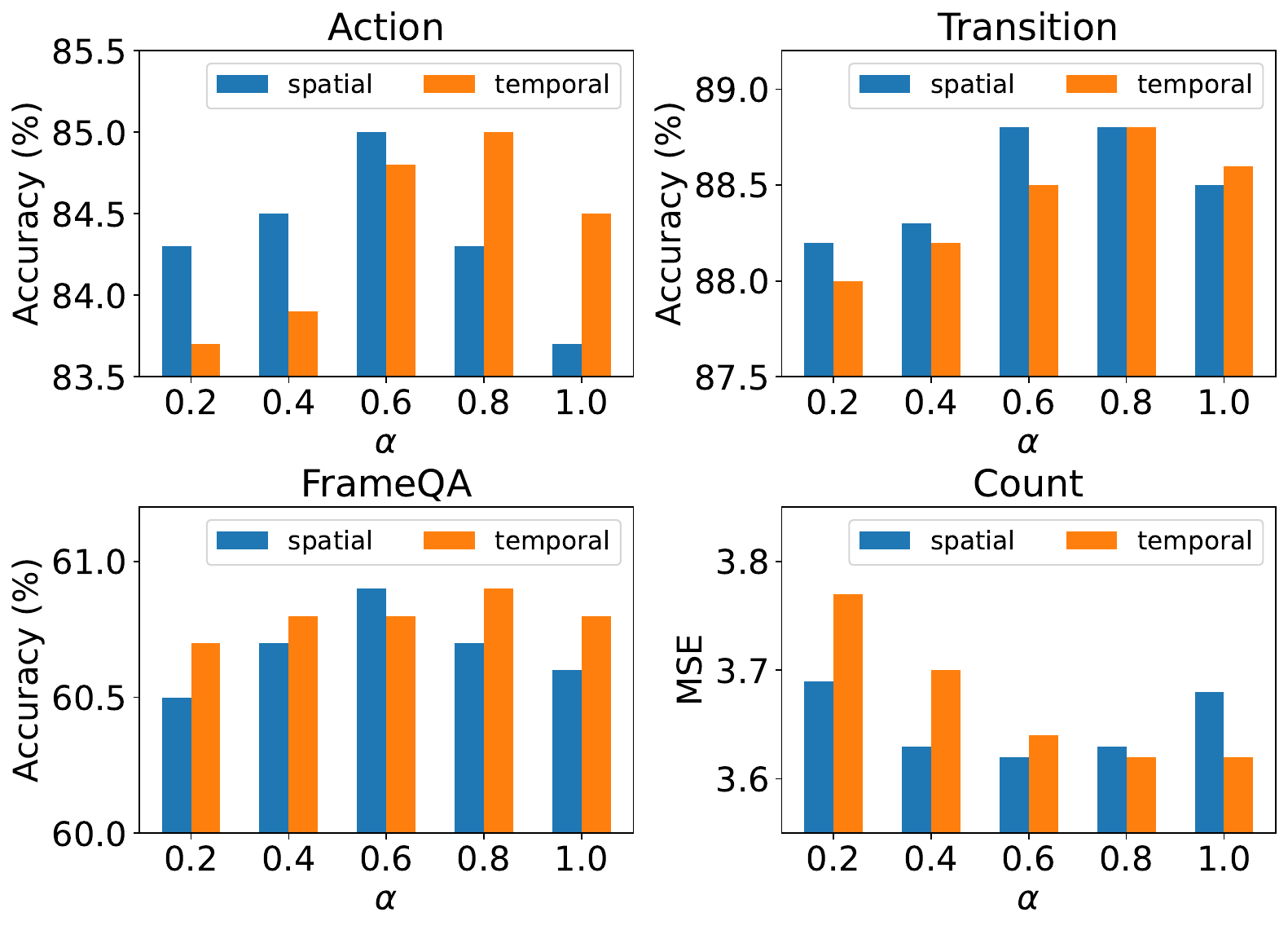}
\end{center}
\caption{Impact of the ratio $\alpha$ of neighbors for graph local relation modeling. 
The blue (orange) bar labelled as spatial (temporal) indicates changing $\alpha$ in spatial (temporal) graphs and fixing $\alpha$ in temporal (spatial) graphs.
The lower the better for \textit{Count}.
}
\label{fig:para1}
\end{figure}


\begin{table}
\renewcommand\arraystretch{1.1}
\caption{Ablation study for different pooling methods.
The lower the better for \textit{Count}. 
}
\begin{center}
\setlength{\tabcolsep}{4.75pt}
\scalebox{0.9}{
    \begin{tabular}{l  c c c | c  c c c }
        \hlinewd{0.4pt}
        \hline
        \multirow{2}{*}{Graphs} & \multicolumn{3}{c|}{Pooling}  & \multirow{2}{*}{Action$\uparrow$} & \multirow{2}{*}{Transition$\uparrow$} & \multirow{2}{*}{Frame$\uparrow$} & \multirow{2}{*}{Count$\downarrow$} \\ \cline{2-4}
          & Max  & Mean  & Sum  &  \\ 
        \hline \hline
        & \checkmark &            &           & \textbf{85.0} & \textbf{88.8}  & \textbf{60.9}   & \textbf{3.62}  \\ 
        spatial &    &\checkmark  &                 & 84.1 & 88.4  & 60.4   & 3.73   \\
           &       &            & \checkmark  & 84.0 & 88.0  & 60.3   & 3.72  \\ 
        \hline
        & \checkmark &            &           & \textbf{85.0} & \textbf{88.8}  & \textbf{60.9}   & \textbf{3.62}  \\
        aggregation  &  &\checkmark  &                  & 84.7 & 88.6  & 60.8   & 3.65   \\
        &          &            & \checkmark  & 84.9 & 88.7  & 60.7   & 3.66  \\ 
        \hline
        & \checkmark &            &           & 84.7 & 88.7  & 60.7   & 3.67 \\
        temporal  &  &\checkmark  &                  & 84.5 & 88.6  & 60.6   & 3.70   \\
        &          &            & \checkmark  & \textbf{85.0} & \textbf{88.8}  & \textbf{60.9}   & \textbf{3.62}  \\ 
        \hline
    \hlinewd{0.4pt}
    \end{tabular}}
\end{center}
\label{table:abl_agg}
\end{table} 

\begin{figure}[t]
\begin{center}
  \includegraphics[width=1.0\linewidth]{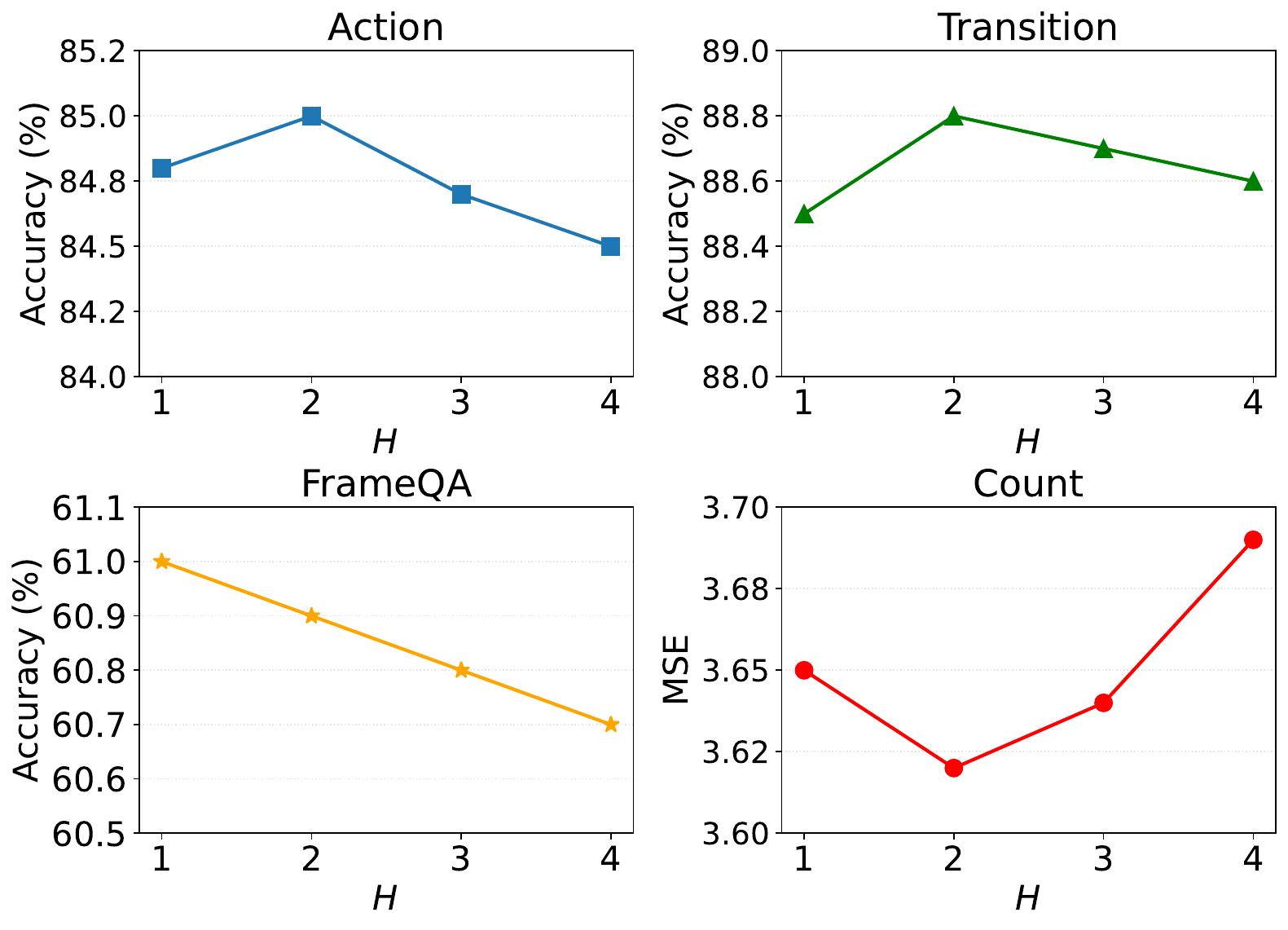}
\end{center}
\vspace{-1em}
\caption{Impact of the number of graph layers $H$.
The lower the better for \textit{Count}.}
\label{fig:para2}
\end{figure}
\vspace{-2em}

\subsection{Ablation Study}
\label{Sec:analysis}
In this section, we conduct a set of ablation studies on TGIF-QA dataset. First, we study the impact of important hyper-parameters (number of neighbors in graphs and number of graph layers). Then, we compare different pooling methods in graph construction. Lastly, we verify the effectiveness of each proposed component.

\paragraph{Number of neighbors in graphs} 
The hyper-parameter $\alpha \in [0,1]$ is a ratio to control the number of neighboring nodes in graphs.
A higher value of $\alpha$ means information from more neighboring nodes is aggregated to compute the center node. To study the impact of $\alpha$, we train the model using different values of $\alpha$. Note that we fix $\alpha$  as $0.6$ ($0.8$) in spatial (temporal) graphs when changing $\alpha$ in the temporal (spatial) graphs. 
As shown in Fig.~\ref{fig:para1}, the hyper-parameter $\alpha$ has significant impact on model performance, and both a lower and a higher $\alpha$ may lead to a performance drop. The reason could be that a higher $\alpha$ may bring much noise from irrelevant objects, while a lower $\alpha$ may miss necessary relation information. Furthermore, there may exist redundant object proposals with poor quality, so $\alpha$ in spatial graphs should be relatively lower than that in temporal graphs. 

\begin{table}
\renewcommand\arraystretch{1.1}
\caption{Ablation study for each proposed component. 
}
\begin{center}
\scalebox{1.0}{
    \begin{tabular}{l c c c c }
    \hlinewd{0.95pt}
         Model & Action$\uparrow$ & Transition$\uparrow$ & Frame$\uparrow$ & Count$\downarrow$  \\
        \hline \hline
        \textbf{Keyword Attention} \\
        \quad w/o Word Attention       &   84.2    &   87.9    &   60.3    &   3.70  \\
        \quad w/o Object Attention             &   83.6    &   87.0    &   59.7    &   3.81  \\
        \hline
        \textbf{Relative Reasoning } \\
         \quad w/o Relative Relation   & 83.8  &   87.2    &   59.8    &   3.70     \\
         \quad w/o Absolute Relation   & 84.0  &   87.1    &   60.1    &   3.72     \\
        \hline
        \textbf{Disentangled Graphs} \\
        \quad w/o Disentangling    &   83.3    &   86.7    &   60.2    &   3.76  \\
        \hline
        \textbf{KRST full model}   &\textbf{85.0}   &\textbf{88.8}  & \textbf{60.9}  &  \textbf{3.62}  \\
        \hline 
    \hlinewd{0.8pt}
    \end{tabular}}
\end{center}
\label{table:abl_comp}
\end{table}

\begin{figure*}[t]
\begin{center}
  \includegraphics[width=0.9\linewidth]{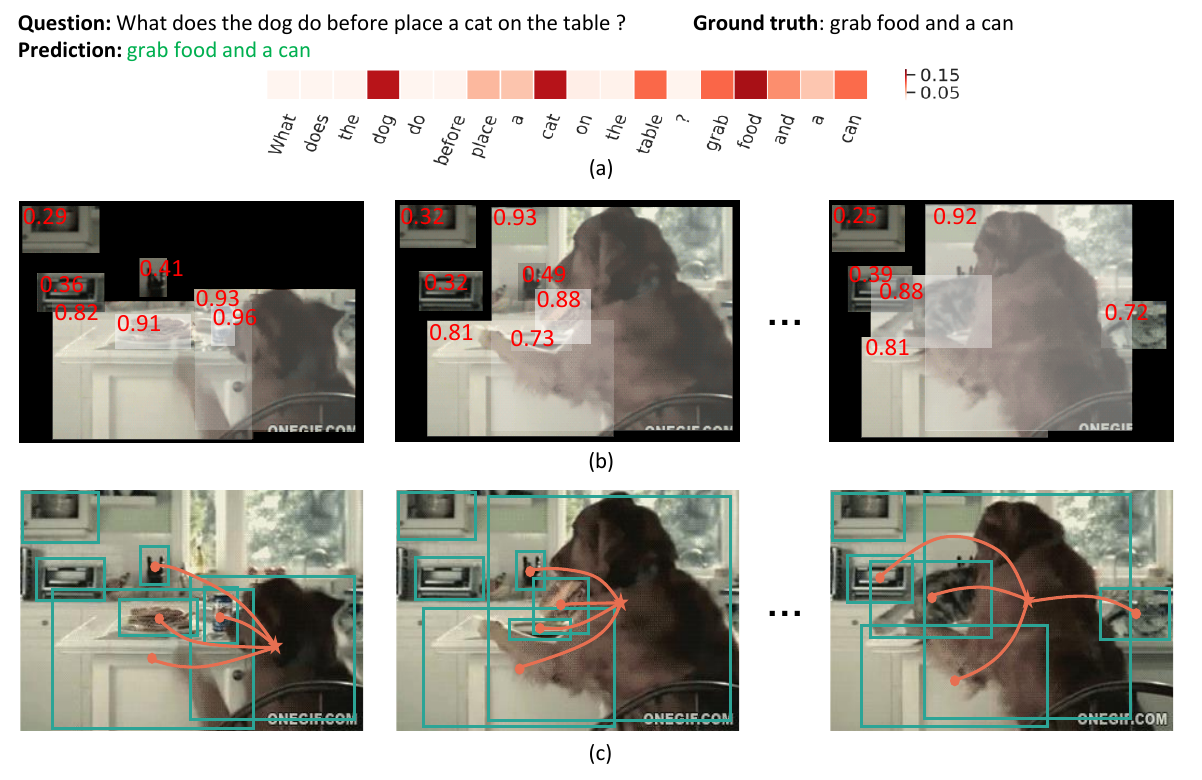}
\end{center}
\caption{
Visualization of (a) word attention $\mA_w$, (b) object attention $\mA_o$ and (c) the k-nearest ($k=4$) relevant neighbors of the dog according to the question. 
In (a) and (b), our method pays high attention to the keywords and the relevant objects, such as ``dog'', ``cat'', \etc. 
In (c), our method selects the most relevant objects as the neighbors in the representation space for graph reasoning. 
}
\label{fig:vis}
\end{figure*}

\paragraph{Number of graph layers $H$} 
We train our model using different values of $H$ and compare the performance in Fig.~\ref{fig:para2}. It is observed that model performance on different tasks reacts differently towards the increase of $H$. On \textit{FrameQA} task, the model achieves the best performance when $H=1$ and starts to drop as $H$ continues increasing. This suggests that a one-layer graph is sufficient for \textit{FrameQA} task, because questions in this task focus on spatial relations and generally do not require complex relation reasoning. On the other tasks, the model achieves the best performance when $H=2$, starting to drop as $H$ continues increasing. This suggests that one-layer graph may be insufficient to perform complex object relation reasoning, while a deep graph model may lead to the over-smoothing problem, resulting in inferior performance. 

\begin{figure*}[t]
\begin{center}
  \includegraphics[width=0.86\linewidth]{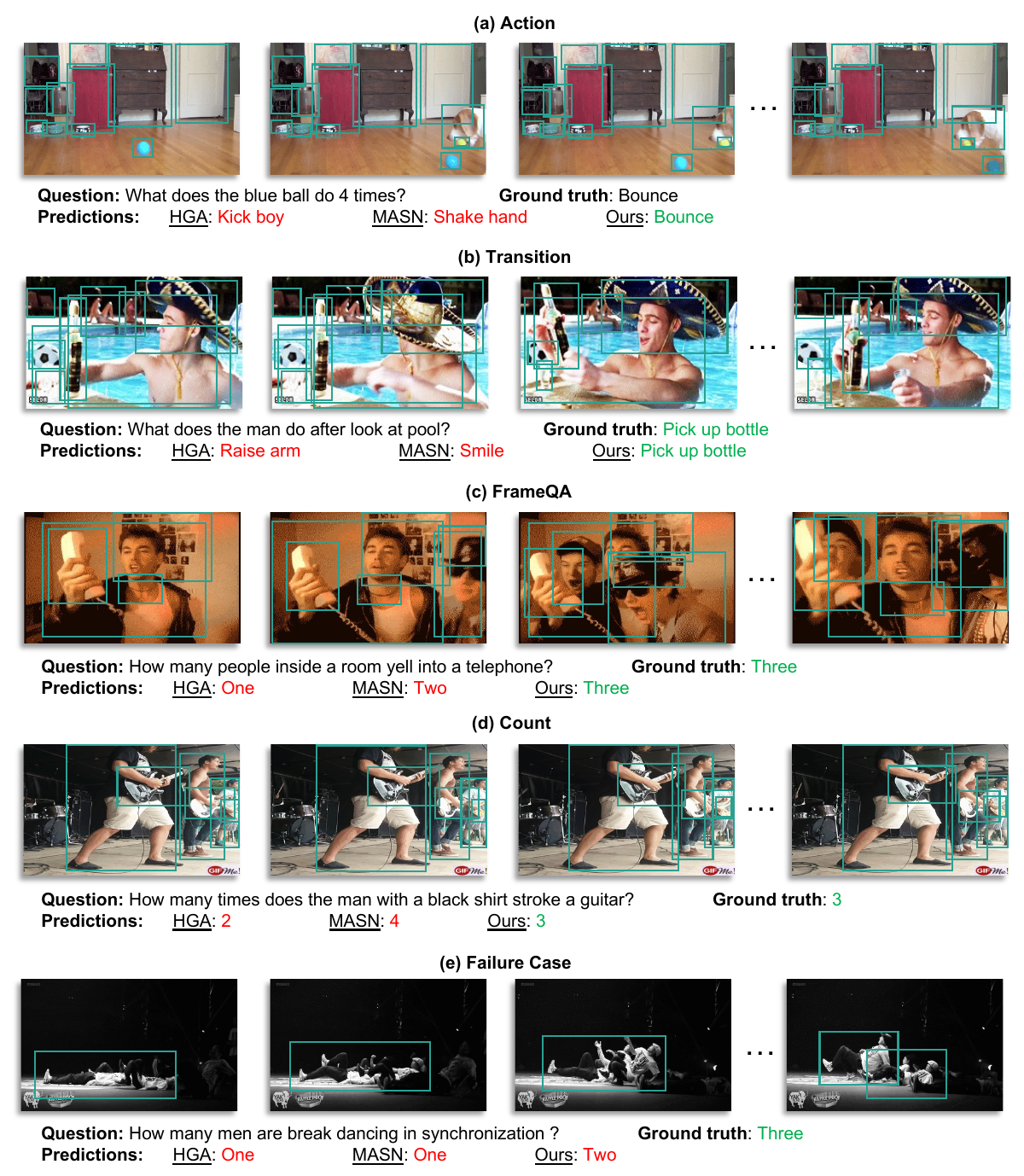}
\end{center}
\caption{Examples from the TGIF-QA dataset. The examples cover four different tasks: Action, Transition, FrameQA and Count. Correct answers (Ground truths) are shown in green and wrong predictions are in red. Rectangles denote the detected object bounding boxes. }
\label{fig:vis_sota}
\end{figure*}

\paragraph{Impact of pooling methods}
We explore different pooling methods in Eq.~\ref{eq:aggr} and summarize the results in Table~\ref{table:abl_agg}. It is observed that maximum pooling generates the best performance for spatial graphs while sum pooling generates the best performance for temporal graphs. As for aggregation, the three pooling methods have similar performance, with mean pooling showing the best result.

\paragraph{Effectiveness of each proposed component} 
To verify the effectiveness of each component in the proposed method, we train ablation models under different settings and summarize the results in Table~\ref{table:abl_comp}. 
Overall, we find that removing any component in KRST would reduce the model performance. 
The impact of each component is detailed as follows.

\textbf{Keyword attention} mainly includes word attention $\mA_w$ and object attention $\mA_o$. $\mA_w$ is used to generate the attended question feature $\hat{\bm{E}}$, and then $\hat{\bm{E}}$ is used to compute $\mA_o$. 
By removing word attention, we replace $\hat{\bm{E}}$ using the global question embedding to compute $\mA_o$. We find that removing word attention will lead to certain performance drops. This is because the model may not effectively identify all the relevant objects when the keywords are not highlighted. By dropping object attention (\ie, the Keyword Attention module), we observe a more significant performance drop. This is as expected since treating all the objects equally may reduce the reasoning efficiency.

\textbf{Relative reasoning} aims to augment absolute relation with relative relation graph reasoning. It is observed that when removing either the relative relation or the absolute relation in graph modeling, the model performance will drop by around $1\%$ on \textit{Action}, \textit{Transition} and \textit{FrameQA} tasks. These results demonstrate that both relative and absolute relations are important in modeling object interactions.

\textbf{Disentangled graphs} model the spatial relation and temporal relations using separate graphs. We study the impact of disentangled graphs by constructing a holistic graph over all the detected objects, where the spatial and temporal relations are modeled equally. We observe that without disentanglement, the model performance drops by only $0.7\%$ on \textit{FrameQA} task, but by around $2\%$ on \textit{Action} and \textit{Transition} task. This is because the latter two tasks are more challenging as they require reasoning about both spatial and temporal relations, while a holistic graph lacks the capability to handle such scenarios. This confirms the effectiveness of disentangled graphs in performing complex relation reasoning.

\subsection{Qualitative Results}
We visualize the learned attention maps (word attention $\mA_w$ and object attention $\mA_o$) and the k-nearest neighbors of the object.
In Fig.~\ref{fig:vis} (a), we can observe that the keywords such as ``dog'', ``cat'' and ``food'' are highlighted by higher attention weights. These highlighted keywords can help to identify the question-relevant objects for the subsequent relation reasoning. 
In Fig.~\ref{fig:vis} (b), we present the detected objects with their learned attention weights (red numbers) which are also indicated by the brightness of the object regions. It is shown that the keyword-relevant objects (\eg, ``dog'', ``cat'' and ``food'') have higher attention weights and therefore are much more brighter than the keyword-irrelevant objects (\eg, ``closet'', ``oven'' and ``jar''). This implies that the most keyword-relevant objects are effectively identified from the video frames by object attention. 
In Fig.~\ref{fig:vis} (c), we visualize the k-nearest neighbors of the object ``dog''. It is observed that our model can correctly select the most relevant objects as neighbors, which help improve the efficacy of relation reasoning.

We also compare our method with the existing SOTA methods on different tasks of the TGIF-QA dataset in Fig.~\ref{fig:vis_sota}. 
The comparing methods include two graph-based methods: 1) HGA~\cite{jiang2020r} on frame relations and 2) MASN~\cite{seo2021masn} on object relations. 
It is observed that our model can correctly answer the questions based on given videos, which demonstrates the efficacy of our model in object relation reasoning.
In Fig.~\ref{fig:vis_sota} (a), our model can attend to the relevant object (\ie,``blue ball'') from other distracting objects (\eg, ``green ball'') and generate the correct answer. This validates the effectiveness of the keyword attention module in filtering out the question-irrelevant features from videos.
In Fig.~\ref{fig:vis_sota} (b), the example shows that our model can correctly predict the answer to the question, which requires the modeling of both spatial relations (\eg, ``pick up bottle'') and temporal dynamics (\eg, ``after''). 
In Fig.~\ref{fig:vis_sota} (c), our model correctly answers the question by aggregating the information in multiple frames. The reason could be that we apply max pooling to aggregate information from neighboring frames in the temporal graph to maintain critical spatial relations in each frame. 
In Fig.~\ref{fig:vis_sota} (d), our model generates the correct answer by attending to the question-relevant object (\ie, ``man with a black shirt''). This again demonstrates the effectiveness of the proposed keyword attention module. 
In Fig.~\ref{fig:vis_sota} (e), all the three models generate incorrect predictions for this sample. The primary cause could be the object detector's inability to identify all the individuals presenting in the video, due to the heavy occlusions and poor lighting conditions.

\section{Conclusion}
In this paper, we present a Keyword-aware Relative Spatio-Temporal (KRST) graph network for VideoQA. 
Specifically, we apply attention mechanism to generate a keyword-aware question embedding for the construction of video graphs.
To better capture the spatio-temporal relation among object nodes, we introduce relative relation modeling into graph networks.
Furthermore, to explicitly model the spatial and temporal relations, we disentangle the holistic spatio-temporal graph into a spatial graph over objects and a temporal graph over frames. 
Extensive experiments on three datasets demonstrate the effectiveness of our proposed method in performing complex object relation reasoning for VideoQA. In future work, we plan to exploit the hierarchical structure of questions and videos, which may improve the relation reasoning by building fine-grained correspondences between linguistic and visual elements.

\bibliographystyle{IEEEtran}
\bibliography{ref}



 





\end{document}